\documentclass[lettersize,journal]{IEEEtran}

\usepackage{cite}
\usepackage{amsmath,amssymb,amsfonts}
\usepackage{algorithmic}
\usepackage{graphicx}
\usepackage{textcomp}
\usepackage{xcolor}
\usepackage{wrapfig}
\usepackage{tikz}
\usepackage[dvipsnames]{xcolor}
\usepackage{array}
\usepackage[caption=false,font=normalsize,labelfont=sf,textfont=sf]{subfig}
\usepackage{stfloats}
\usepackage{url}
\usepackage{verbatim}
\def\BibTeX{{\rm B\kern-.05em{\sc i\kern-.025em b}\kern-.08em
    T\kern-.1667em\lower.7ex\hbox{E}\kern-.125emX}}
\usepackage{balance}
\usepackage{orcidlink}

\def\BibTeX{{\rm B\kern-.05em{\sc i\kern-.025em b}\kern-.08em
    T\kern-.1667em\lower.7ex\hbox{E}\kern-.125emX}}
\begin{document}

\title{RAGE-XY: RADAR-Aided Longitudinal and Lateral Forces \\Estimation For Autonomous Race Cars}

\author{
    \IEEEauthorblockN{Davide Malvezzi\orcidlink{0009-0003-7603-2025},}
    \IEEEauthorblockN{Nicola Musiu\orcidlink{0009-0001-6832-0997},}
    \IEEEauthorblockN{Eugenio Mascaro,}
    \IEEEauthorblockN{Francesco Iacovacci,}
    \IEEEauthorblockN{and Marko Bertogna\orcidlink{0000-0003-2115-4853}}
}

\maketitle

\begin{abstract}
In this work, we present RAGE-XY, an extended version of RAGE, a real-time estimation framework that simultaneously infers vehicle velocity, tire slip angles, and the forces acting on the vehicle using only standard onboard sensors such as IMUs and RADARs. Compared to the original formulation, the proposed method incorporates an online RADAR calibration module, improving the accuracy of lateral velocity estimation in the presence of sensor misalignment. Furthermore, we extend the underlying vehicle model from a single-track approximation to a tricycle model, enabling the estimation of rear longitudinal tire forces in addition to lateral dynamics. We validate the proposed approach through both high-fidelity simulations and real-world experiments conducted on the EAV-24 autonomous race car, demonstrating improved accuracy and robustness in estimating both lateral and longitudinal vehicle dynamics.

\end{abstract}


\section{Introduction}

In this work, we present RAGE-XY, an extended
version of RAGE \cite{RAGE} (RADAR-Aided Grip Estimator), a real-time estimation framework that jointly estimates the vehicle velocity vector, tire slip angles, and tire forces using only standard onboard sensors, namely an IMU and multiple RADAR units. Compared to the original formulation, this work introduces two key contributions.

First, we incorporate an online RADAR calibration module that compensates for sensor rotation misalignment, significantly improving the accuracy of lateral velocity estimation. This is particularly important in practical deployments, where installation imperfections and mechanical tolerances can degrade performance if left unaccounted for.

Second, we extend the underlying vehicle model from a single-track representation to a tricycle model. This richer modeling allows us to capture longitudinal force dynamics at the rear axle, enabling the joint estimation of both lateral and longitudinal tire forces. As a result, the proposed framework provides a more complete representation of the vehicle–tire interaction, which is crucial in aggressive driving conditions.

\begin{figure}[t]
\centering
{\includegraphics[width=\linewidth,trim={10cm 2cm 0 11cm},clip]{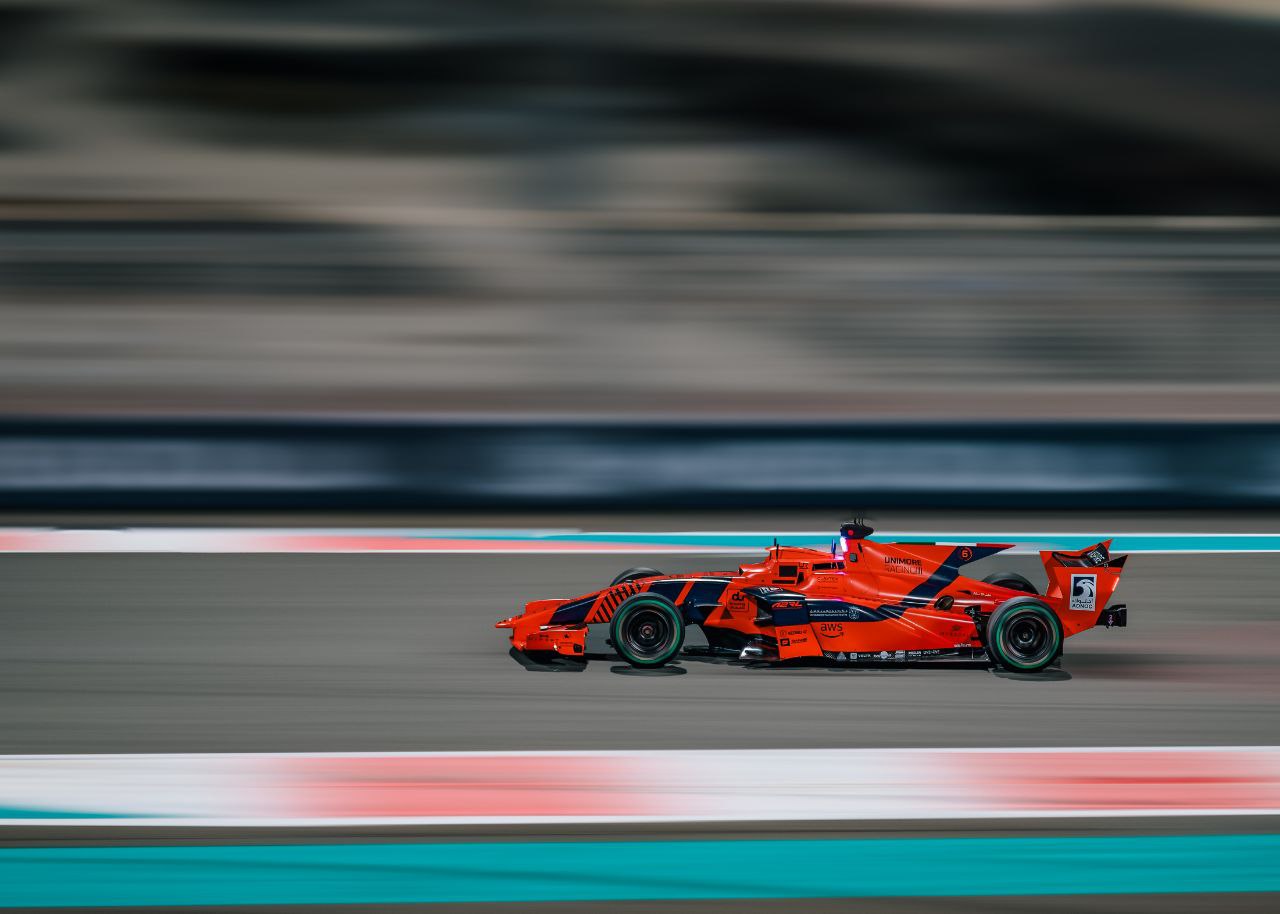}
\caption{Unimore Racing’s Dallara EAV-24 during the 2024 Abu Dhabi Autonomous Racing League at Yas Marina F1 circuit.}}
\label{fig:car}
\end{figure}

The proposed framework is first validated in a high-fidelity simulation environment based on a multi-body vehicle model \cite{ER_AUTOPILOT}, enabling controlled evaluation under extreme driving conditions. It is then assessed on real-world data collected from the EAV-24 autonomous race car (see Figure \ref{fig:car}) during the Abu Dhabi Autonomous Racing League (A2RL\footnote{www.a2rl.io
}), where the Unimore Racing Team\footnote{www.unimoreracing.com
} competed. The dataset includes high-speed scenarios with velocities up to 70 m/s and lateral accelerations up to 28 $\text{m/s}^2$, providing a comprehensive benchmark to evaluate the accuracy, robustness, and adaptability of the proposed method across a wide range of operating conditions.
\section{Related Work} \label{sec:related_works}
Direct measurement of lateral velocity and slip angles typically requires expensive instrumentation, such as optical sensors or high-precision inertial navigation systems, which limits their applicability outside controlled environments. As a result, model-based estimation techniques relying on standard onboard sensors have become widely adopted.
Early approaches are based on kinematic models \cite{KINEMATIC_MODEL}, while more advanced methods combine kinematic and dynamic formulations \cite{DYNAMIC_KINEMATIC_MODEL}, improving estimation accuracy under aggressive driving conditions. Extensions incorporating tire models, such as the Dugoff \cite{DYNAMIC_KINEMATIC_MODEL2} or Pacejka formulations \cite{MODEL_BASED}, enable the estimation of lateral tire forces, although they typically rely on pre-identified parameters that do not adapt to changing operating conditions.

Data-driven approaches have also been explored, with neural network-based methods achieving strong performance in racing scenarios \cite{NN1} and demonstrating robustness to tire wear and varying grip conditions \cite{NN2}. However, these methods require extensive training data and often exhibit limited generalization when deployed in previously unseen conditions \cite{NN3}.

Several works exploit additional sensing modalities. GNSS-aided methods \cite{GNSS_SLIP, GNSS_SLIP2, GNSS_SLIP3, GNSS_SLIP4, GNSS_DOUBLE_ANTENNA_SLIP} combine IMU and satellite measurements to estimate slip angles, but suffer from limited update rates and degraded performance in GNSS-denied environments. Vision-based approaches \cite{CAMERA_SLIP_DOWNFACE, CAMERA_SLIP_DOWNFACE2, CAMERA_SLIP_DOWNFACE3, CAMERA_SLIP_CORRELATION, CAMERA_SLIP_RC_SCALED} estimate ground-relative motion using downward-facing cameras, though they require textured surfaces, precise calibration, and high frame rates at elevated speeds. Similarly, LiDAR-based methods, such as our previous LOP-UKF framework \cite{LOP_UKF}, rely on accurate localization against pre-built maps and offline-calibrated tire parameters.

Several observer-based approaches have been proposed for longitudinal tire force estimation. Ultra-local modeling techniques \cite{FX_ULM} estimate longitudinal forces with minimal modeling assumptions, but require lateral force inputs, limiting their standalone applicability. Nonlinear observers, such as sliding-mode approaches \cite{FX_NON_LINEAR_OBS}, reduce sensor requirements but rely on prior knowledge of tire–road friction coefficients and tire model parameters. Other formulations, including LPV-based methods \cite{FX_LPV}, aim to capture longitudinal force characteristics through simplified parameterizations. However, these approaches often require force measurements as inputs.

More comprehensive frameworks attempt to jointly estimate longitudinal and lateral dynamics. For instance, T.R.I.C.K 2.0 \cite{TRICK2} models full vehicle dynamics with high fidelity, but depends on an extensive sensor suite, increasing system complexity and limiting practical deployment. Similarly, interacting multiple-model approaches \cite{FX_FY_IMM} have been proposed, particularly for electric vehicles, where longitudinal and lateral dynamics are handled through switching mechanisms.

A common alternative is represented by hierarchical or cascaded estimation strategies, where the problem is decomposed into multiple stages. In \cite{CASCADED_FY_THEN_FX}, an adaptive sliding-mode observer is first used to estimate the four longitudinal tire forces independently. Lateral forces are then computed using a simplified Dugoff-based formulation, and finally, vehicle states are estimated. Similarly, in \cite{FX_THEN_FY}, a proportional–multiple–integral observer based on an LPV formulation is proposed, with parameters tuned via particle swarm optimization. The method follows a sequential strategy: longitudinal tire forces are first estimated and subsequently treated as known inputs for lateral force estimation. Other multi-stage approaches include \cite{PARAMS_THEN_FORCES_THEN_STATE}, where vehicle mass is first estimated via recursive least squares, followed by an adaptive sliding-mode observer for tire force estimation, and finally state estimation. In \cite{STATE_THEN_FZ_THEN_FX_FY}, vertical forces are first computed considering load transfer effects, and longitudinal and lateral forces are subsequently estimated using a sliding-mode observer.

While these hierarchical approaches can achieve good performance, their staged structure introduces strong interdependencies between estimation blocks, making them sensitive to modeling errors and uncertainty propagation.

The proposed estimator builds upon these observations and introduces the following key contributions:

\begin{itemize}
\item a unified estimation framework that jointly infers the full vehicle state together with lateral and longitudinal tire forces in real time, avoiding the limitations of sequential estimation schemes;

\item the use of multiple RADAR sensors in combination with an IMU to reconstruct both longitudinal and lateral velocity components without relying on high-cost sensors;

\item an online RADAR calibration procedure that compensates for sensor misalignment, improving the accuracy and consistency of velocity estimation;

\item online identification of lateral tire model parameters, enabling adaptation to changing road conditions, tire wear, and thermal effects;

\item a lightweight model of a limited-slip differential, allowing a more accurate representation of rear-axle force distribution.

\end{itemize}

\section{Preliminaries}\label{sec:preliminaries}
In this paper, matrices are written in bold uppercase letters (e.g. $\mathbf{R}$), while vectors are written as bold lowercase letters (e.g. \textbf{v}). A rigid-body transformation from frame $B$ to frame $A$ is represented as $(\mathbf{R}^A_B, \textbf{t}^A_B) \in \text{SO(3)} \times \mathbb{R}^3$ or with its corresponding transformation matrix $\textbf{T}^A_B \in \text{SE(3)}$. The velocity of a frame $B$ expressed with respect to a frame $A$ is denoted by $\mathbf{v}_B^A$.  The body frame, denoted by $B$, is defined at the vehicle’s Center of Gravity (CoG). It is assumed that the IMU reference frame coincides with the body frame. The RADAR reference frame is denoted by $R$. Sensor measurements and any values derived from them are denoted with a hat symbol (e.g. $\boldsymbol{\hat{\omega}}$). 

\section{System Overview} \label{sec:overview}
\begin{figure}
    \centering
    \includegraphics[width=\linewidth]{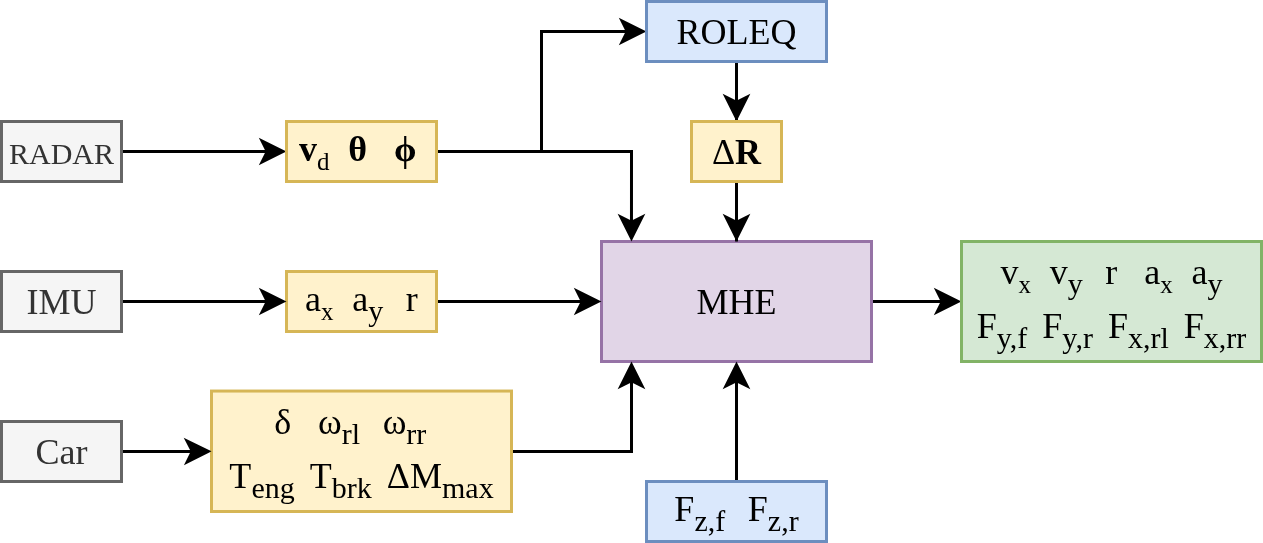}
    \caption{Estimator overview.}
    \label{fig:system}
\end{figure}

An overview of the system architecture, including inputs and outputs, is presented in Fig. \ref{fig:system}. RAGE-XY is formulated as Moving Horizon Estimation (MHE) problems. The estimation objective is to obtain the maximum a posteriori estimate of the ego-vehicle states and tire forces by minimizing a cost function that accounts for residuals associated with prior information, system dynamics, and available measurements. The MHE framework is particularly well suited to handle delayed or asynchronous measurements, as long as their timestamps lie within the active estimation horizon. This capability is especially relevant for RADAR measurements, which may exhibit non-negligible delays due to internal sensor processing.

Concisely, the MHE problem can be defined as follows
\begin{align*}
\underset{\Omega}{\mathrm{min}} \quad
&\|\mathbf{x}_0 - \tilde{\mathbf{x}}_0\|^2_{\mathbf{\Sigma}_{\mathbf{x}_0}} + 
 \|\mathbf{P} - \tilde{\mathbf{P}}\|^2_{\mathbf{\Sigma}_{\mathbf{P}}} \\
&+ \sum_{k=0}^{N} \left( 
    \|\mathbf{w}_k\|^2_{\mathbf{\Sigma}_{\mathbf{w}}} + 
    \sum_{l=0}^{M} \|\mathbf{v}^l_k\|^2_{\mathbf{\Sigma}_{\mathbf{v}^l}} 
\right)
\end{align*}
\begin{align*}
\text{subject to} \\
\mathbf{x}_{k} &= f(\mathbf{x}_{k-1}, \mathbf{u}_{k-1}) + \mathbf{w}_k \qquad \qquad \quad k = 1, \dots, N \\
\mathbf{\hat{y}}^l_k &= h_l(\mathbf{x}, \mathbf{u}, \mathbf{P}) + \mathbf{v}^l_k  \quad k = 0, \dots, N \, l = 0, \dots, M \\
&\mathbf{P}_{\min} \leq \mathbf{P} \leq \mathbf{P}_{\max}
\end{align*}
The set of variables to be estimated is defined as $\boldsymbol{\Omega} = [\boldsymbol{X}, \boldsymbol{P}]$ where $\boldsymbol{X}$ denotes the sequence of states over a sliding time window of length (N+1), and $\boldsymbol{P}$ represents a set of parameters to be estimated.
The MHE optimization problem is solved in a receding-horizon fashion. As the estimation window advances, the initial state is updated to the second state of the previous solution, a new state is appended at the end of the horizon, and the most recent measurements are incorporated while the oldest ones are discarded. The quantity $\tilde{\mathbf{x}}_0$ acts as a prior for the initial state of the current window and is obtained from the corresponding state estimate of the previous iteration, thereby ensuring temporal consistency. The associated covariance matrix $\mathbf{\Sigma}_{\mathbf{x}_0}$ regulates the allowable deviation from this prior, promoting smoothness and continuity in the state evolution. Similarly, $\tilde{\mathbf{P}}$ represents the prior for the parameter vector $\mathbf{P}$, which is updated at each iteration using the parameter estimate obtained in the previous step. The covariance matrix $\mathbf{\Sigma}_{\mathbf{P}}$ determines the rate at which the parameters can adapt to new measurements, effectively balancing the retention of past information against responsiveness to recent data. The function $f(\cdot)$ defines the system state-transition model, predicting the next state $\mathbf{x}_{k+1}$ from the current state $\mathbf{x} $ and the control inputs $\mathbf{u}$. The set of functions $h_l(\cdot)_{l=0}^{M}$ describes the measurement models, which map the states, inputs, and parameters to the corresponding sensor outputs $\hat{\mathbf{y}}^{l}_k$. Both the process and measurement models are assumed to be affected by additive zero-mean Gaussian noise, namely process noise $\mathbf{w}_k \sim \mathcal{N}(0, \mathbf{\Sigma}_{\mathbf{w}})$ and measurement noise $ \mathbf{v}^{l}_k \sim \mathcal{N}(0,\mathbf{\Sigma}_{\mathbf{v}^{l}})$. Finally, the parameters $ \mathbf{P} $ are constrained within predefined bounds to ensure physical plausibility.

\section{Lateral Forces Estimation} \label{sec:lateral}
RAGE-XY estimates both the vehicle states and the lateral forces acting on the vehicle. The state vector includes the longitudinal and lateral velocities $(v_x, v_y)$, longitudinal and lateral accelerations $(a_x, a_y)$, and the yaw rate ($r$). The velocity dynamics are described using a simple point-mass model, the accelerations are modeled with a constant-jerk assumption, and the yaw rate follows a constant-angular-acceleration model.
To account for drift in the IMU measurements, the corresponding sensor biases $(b_x, b_y, b_r)$ are also included in the state vector and estimated online. These biases are modeled as random-walk processes, allowing them to evolve gradually over time and enabling the estimator to correct for long-term sensor offsets.

\begin{figure}[b]
  \centering
  \begin{tikzpicture}[scale=1.7, every node/.style={font=\small}]

    \draw[thick] (2, 2) -- (5, 2) node[midway, above] {};

    \draw[dashed] (2-0.4, 2-0.2) rectangle (2+0.4, 2+0.2);
    \node[below] at (2, 3.3) {Rear axle};

    \draw[thick] (2-0.4, 2.8-0.2) rectangle (2+0.4, 2.8+0.2);
    \draw[thick] (2-0.4, 1.2-0.2) rectangle (2+0.4, 1.2+0.2);
    \draw[thick] (2, 1.2) -- (2, 2.8) node[midway, above] {};

    \draw[->, thick, blue] (2, 2.8) -- (2.7, 2.8) node[midway, above right] {$F_{x,rl}$}; 
    \draw[->, thick, blue] (2, 1.2) -- (2.7, 1.2) node[midway, above right] {$F_{x,rr}$}; 
    \draw[->, thick, red] (2, 2) -- (2, 2.7) node[pos=0.3, above left] {$F_{y,r}$};  
    \draw[->, thick, black] (2, 2) -- (2.6, 1.85);
    \draw[->] (2, 2) ++(0:0.5) arc[start angle=0, end angle=-14, radius=0.5]; 
    \node[right] at (2.55, 1.92) {$\alpha_r$}; 

    \begin{scope}[shift={(5, 2)}] 
        \begin{scope}[rotate=35]
            \draw[thick] (-0.4, -0.2) rectangle (0.4, 0.2);
            \draw[->, thick, red] (0, 0) -- (0, 0.7) node[pos=0.7, above right, right=0] {$F_{y,f}$};  

            \draw[ultra thin, dashed] (0, 0) -- (1.1, 0);

            \draw[->, thick, black] (0, 0) -- (0.6, -0.15);
            \draw[->] (0.0, 0) ++(0:0.5) arc[start angle=0, end angle=-14, radius=0.5]; 
            \node[right] at (0.55, -0.04) {$\alpha_f$}; 
        \end{scope}

        \draw[->] (0.5, 0) ++(0:0.5) arc[start angle=0, end angle=46, radius=0.75]; 
        \node[right] at (0.95, 0.35) {$\delta$}; 
        \draw[ultra thin, dashed] (0, 0) -- (1.1, 0);
    \end{scope}
    \node[below] at (5, 3.3) {Front axle};


    \fill[orange] (3.5, 2) circle (2pt);
    \node[below] at (3.5, 2.0) {CoG};

    \draw[<->] (2, 0.9) -- (3.5, 0.9) node[midway, below] {$l_r$}; 
    \draw[<->] (3.5, 0.9) -- (5, 0.9) node[midway, below] {$l_f$}; 
    \draw[<->] (1.5, 1.2) -- (1.5, 2.8) node[midway, below,xshift=-0.2cm, yshift=0.2cm] {$t_r$}; 

    \begin{scope}[shift={(3.5, 2)}] 
    \draw[->, thick] (0, 0) -- (0.7, 0.35) node[anchor=south] {$\vec{v}$};
    \draw[->, thick] (0, 0) -- (0.7, 0) node[anchor=north] {$v_x$};
    \draw[->, thick] (0, 0) -- (0, 0.35) node[anchor=south] {$v_y$};
    \draw[->] (0, 0) ++(0:0.5) arc[start angle=0, end angle=26, radius=0.5]; 
        \node[right] at (0.5, 0.15) {$\beta$}; 
    \end{scope}

    \draw[<-, thick] (3.0, 2.4) ++(0.3, 0.3) arc[start angle=120, end angle=60, radius=0.5];
    \node[above right] at (3.4, 2.8) {$r$}; 
  \end{tikzpicture}
  \caption{Schematic representation of the hybrid tricycle model. For lateral force estimation, an equivalent single-track model is adopted, allowing the estimation of the lateral forces acting at the front and rear axles. For longitudinal force estimation, a tricycle model is considered instead, enabling the estimation of the longitudinal forces acting on the two rear wheels.}
\label{fig:bicycle}
\end{figure}
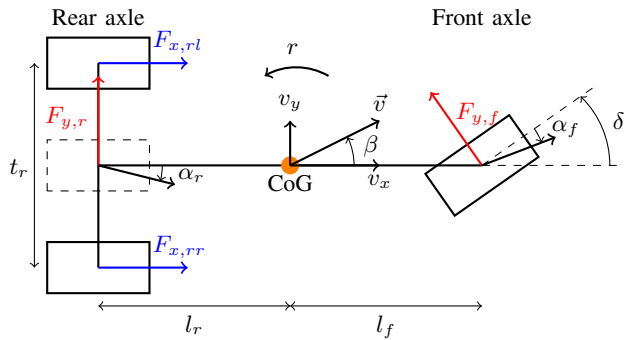

The overall dynamics are then the following
\begin{align*}
    \dot{v}_{x} &= a_x + r \, v_{y} &\dot{v}_{y} &= a_y - r \, v_{x} \\
    \dot{a}_{x} &= 0  &\dot{b}_{x} &= 0 \\
    \dot{a}_{y} &= 0   &\dot{b}_{y} &= 0 \\
    \dot{r} &= 0   &\dot{b}_{r} &= 0
\end{align*}
For the estimation of lateral tire forces, a single-track vehicle model (Fig. \ref{fig:bicycle}) is adopted. Within this modeling framework, only the total lateral forces acting at the front and rear axles can be estimated. The relationship between tire slip angles, the vertical load on each axle, and the resulting lateral forces is described using the Pacejka Magic Formula \cite{PACEJKA}. The set of macro-parameters defining the Pacejka tire model for both the front and rear axles is denoted by $\boldsymbol{P}$.
\begin{align*}
    &\boldsymbol{P} = [ \mathbf{p}_{f}, \mathbf{p}_{r} ] \\
    &\mathbf{p}_{j} = [ B_{j}, C_{j}, D_{j}, E_{j}, S_{hj}, S_{vj} ], \quad j \in [f, r]
\end{align*}
where the subscript $j$ indicates whether the parameters apply to the front ($f$) or rear ($r$) axle.

The measurements vector at each time step is:
\[
    \mathbf{m} = [\hat{a}_{x}, \hat{a}_{y}, \hat{r}, \hat{\delta}, \hat{\boldsymbol{v}}_d, \hat{\boldsymbol{\theta}}, \hat{\boldsymbol{\phi}}]
\]
These correspond to the longitudinal acceleration $\hat{a}_x$, the lateral acceleration $\hat{a}_y$ and the yaw rate $\hat{r}$ measured by the IMU, the angle of the steering wheel $\hat{\delta}$ obtained from the vehicle actuation feedback and the Doppler velocity $\hat{\boldsymbol{v}}_d$, azimuth angles $\hat{\boldsymbol{\theta}}$ and elevation angles $\hat{\boldsymbol{\phi}}$  for each of the RADAR detections.

\subsection{Inputs Reconstruction}
In our new formulation, the vehicle accelerations $(a_x, a_y)$ and yaw rate $(r)$ are treated as model inputs and simultaneously estimated within the MHE framework. This approach accounts for uncertainties in the IMU measurements and allows the estimator to filter out measurement noise as well as external disturbances, such as uneven road surfaces, curbs, or other transient effects, improving the robustness and accuracy of the state and lateral force estimates.
The inputs measuraments model is then the following
\begin{align} 
    \hat{a}_x &= a_{x} + b_{x} \\
    \hat{a}_y &= a_{y}  + b_{y} \\
    \hat{r}  &=  r + b_{r}
\end{align}

\subsection{Online Calibration}
The estimated vehicle velocity is highly sensitive to the orientation of the RADAR sensor. Accurate knowledge of this rotation is required to correctly decompose the RADAR-measured velocity into the vehicle’s longitudinal and lateral components. In particular, at high longitudinal speeds, even small errors in the rotation about the z-axis can lead to significant overestimation of the lateral velocity, as part of the longitudinal component is erroneously projected onto the lateral direction.
To compensate for calibration inaccuracies and potential misalignment caused by vibrations, a ROLEQ estimator \cite{ROLEQ} is employed to estimate online the current RADAR rotation during straight-line driving. Under these conditions, the vehicle velocity is assumed to lie entirely along the longitudinal direction $v_x$, while the lateral and vertical components $v_y$ and $v_z$, as well as the yaw rate $r$, are expected to be negligible. Consequently, the relationship between the RADAR velocity and the body velocity simplifies to $\mathbf{v}_B^B = \mathbf{R}_R^{B} \mathbf{v}_R^R$.
By stacking $\hat{v}_d = -\mathbf{b}(\hat{\theta}, \hat{\phi})^{\top} \mathbf{v}_R^R$ for all detections within a single RADAR scan, the vehicle velocity in the RADAR frame $\mathbf{v}_R^R$ can be estimated by solving the resulting linear system. Using this estimate of the RADAR velocity the ROLEQ estimator determines the optimal rotation correction $\Delta \mathbf{R}$ that satisfies
\begin{align}
\begin{bmatrix}
1 \\ 0 \\ 0
\end{bmatrix}=
\Delta \mathbf{R} \mathbf{R}_R^{B} 
\frac{\mathbf{v}_R^R}{\|\mathbf{v}_R^R\|}
\end{align}
The RADAR rotation is then updated by applying the computed correction  as
\begin{align}    
\mathbf{R}_R^{B} \xleftarrow{} \Delta \mathbf{R} \mathbf{R}_R^{B}
\end{align}
Given this formulation, only the yaw and pitch angles of the RADAR with respect to the vehicle frame are observable. Consequently, the roll angle cannot be identified from this measurement setup, since rotations about the longitudinal axis do not affect the alignment of the normalized velocity vector with the reference direction. 

\section{Longitudinal Forces Estimation}

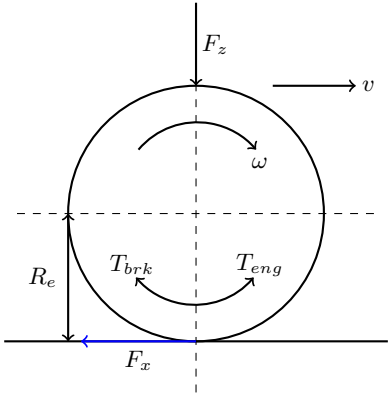
\begin{figure}[tb]
\centering
\begin{tikzpicture}[scale=0.85, every node/.style={font=\small}]

\def\R{2}

\draw[thick] (-3,-\R) -- (3,-\R);

\draw[thick] (0,0) circle (\R);

\draw[dashed] (-2.8,0) -- (2.8,0);
\draw[dashed] (0,-2.8) -- (0,2.8);

\draw[->, thick] (1.2,2) -- (2.5,2);
\node at (2.7,2) {$v$};

\draw[->, thick] (-0.9,1.0) arc (140:40:1.2);
\node at (1.0,0.8) {$\omega$};

\draw[<->, thick] (0.9,-1) arc (-40:-140:1.2);
\node at (1.0,-0.8) {$T_{eng}$};
\node at (-1.0,-0.8) {$T_{brk}$};

\draw[->, thick, blue] (0,-\R) -- (-1.8,-\R);
\node at (-0.9,-\R-0.3) {$F_x$};

\draw[<-, thick] (0,\R) -- (0,\R+1.3);
\node at (0.3,\R+0.65) {$F_z$};

\draw[<->, thick] (-2,0) -- (-2,-\R);
\node at (-2.4,-1) {$R_e$};

\end{tikzpicture}
\caption{Single-wheel model of vehicle.}
\label{fig:single_wheel_model}
\end{figure}

Rather than estimating the individual wheel longitudinal forces directly, they are re-parameterized as
\begin{align}
F_{x,rl} &= \gamma  F_{x,r}, \\
F_{x,rr} &= (1 - \gamma) F_{x,r}
\end{align}
where $F_{x,r}$ denotes the total longitudinal force acting on the rear axle and $\gamma \in [0,1]$ is a split coefficient describing how this total force is distributed between the left and right wheels. This parameterization is motivated by the fact that measurements of the individual wheel forces are not available, whereas an estimate of the total longitudinal force can be inferred from the longitudinal acceleration $a_x$. Moreover, the forces cannot be assumed identical because the race car considered in this study is equipped with a Limited-Slip Differential (LSD), which redistributes engine torque between the wheels when slip occurs.
The estimator state is augmented with the longitudinal dynamics of the vehicle by applying the conservation of angular momentum to both the driven wheels, as illustrated in Fig. ~\ref{fig:single_wheel_model} , and to the vehicle body, while $F_{x,r}$ and $\gamma$ are modeled as random-walk processes
\begin{align}
I_z \dot{r} &= F_{y,f} \cos(\delta) l_f - F_{y,r} l_r + \frac{1}{2}(F_{x,rr} - F_{x,rl}) t_r \\
I_w \dot{\omega}_{rl} &= \frac{T_{eng} - \Delta M}{2} - T_{brk,rl} - F_{x,rl} R_{e,rl} \\
I_w \dot{\omega}_{rr} &= \frac{T_{eng} + \Delta M}{2} - T_{brk,rr} - F_{x,rr} R_{e,rr} \\
\dot{F}_{x,r} &= 0 \\
\dot{\gamma} &= 0 \\
\Delta\dot{M} &= 0
\end{align}
subject to $\gamma \in [0,1]$ and $\Delta M \in [\Delta M_{min}, \Delta M_{max}]$.

In this model, $F_{y,f}$ and $F_{y,r}$ are the estimated lateral forces, $t_r$ is the rear track width, $I_z$ is the vehicle yaw inertia, $I_w$ is the wheel inertia, $T_{eng}$ is the engine torque delivered to the rear axle, $T_{brk,k}$ is the braking torque applied at each wheel, and $R_{e,k}$ denotes the effective rolling radius.

The engine torque $T_{eng}$ is obtained from a fitted map that relates engine speed and throttle command to the delivered torque. 
The braking torque applied at each wheel is modeled as
\begin{align}
T_{brk,k} = \mu_p \, P_k \, A \, r_b
\end{align}
where $\mu_p$ denotes the friction coefficient between the brake pad and the rotor, $P_k$ is the hydraulic pressure in the $k$-th brake circuit, $A$ is the effective caliper piston area, and $r_b$ is the effective radius of the brake disc.

A slack variable $\Delta M$ is introduced to capture the torque redistribution induced by the LSD. Its admissible range is determined using the simplified differential model proposed in \cite{LSD_DIFF_MUSIU}:
\begin{align}
\Delta M_{diff} = \xi \, \max(M_0, \epsilon \, T_{eng}) \, \mathrm{sign}(r)
\end{align}
where $M_0$ represents the preload moment that guarantees a minimum locking effect, $\xi$ is an engagement factor that reduces the locking action during straight-line driving, and $\mathrm{sign}(r)$ identifies the turning direction. The coefficient $\epsilon$ defines the fraction of engine torque that can be transferred between the rear wheels to preserve traction when one wheel loses grip. The value of  $\epsilon$ differs between driving and coasting phases. The bounds on $\Delta M$ are finally defined as
\begin{align}
\Delta M_{min} &= \min(0, \Delta M_{diff}) \\
\Delta M_{max} &= \max(0, \Delta M_{diff})
\end{align}

\subsection{Rear Longitudinal Force }
Measurements of the total rear longitudinal force can be obtained by compensating for the resistive forces acting on the vehicle, modeled according to the tricycle representation shown in Fig.~\ref{fig:bicycle}. These loss forces are modeled as the sum of aerodynamic drag, rolling resistance, cornering force, and road slope effects:
\begin{align}
F_{\text{loss}} &= F_{\text{aero}} + \sum_j F_{\text{roll},j} + F_{\text{corner}} + F_{\text{slope}} \\
F_{\text{aero}} &= \tfrac{1}{2}\, \rho\, c_x\, A\, v_x^{2} \\
F_{\text{roll},j} &= C_r\, \hat{F}_{z,j}\, \exp\!\left(-\frac{2}{\hat{\omega}_j R_{e,j}}\right) \\
F_{\text{corner}} &= F_{y,f}\, \sin(\alpha_f) + F_{y,r}\, \sin(\alpha_r) \\
F_{\text{slope}} &= m g \sin(\alpha_s)
\end{align}
where \(\rho\) is the air density, \(c_x\) is the aerodynamic drag coefficient, \(A\) is the vehicle frontal area, \(C_r\) is the rolling resistance coefficient, \(\hat{F}_{z,j}\) is the measured vertical load on the \(j\)-th wheel, \(\hat{w}_{j}\) is the measured wheel angular velocity of the \(j\)-th wheel,  \(m\) is the vehicle mass, and \(\alpha_s\) denotes the road slope.
The total longitudinal force acting on the rear axle is then computed as
\begin{align}
\hat{F}_{x,r} =
\begin{cases}
\left( m + \displaystyle\sum_j \frac{I_{\omega}}{R_j^{2}} \right) a_x + F_{\text{loss}} & \text{(driving)} \\
(1 - \beta_b)\, \big( m a_x + F_{\text{loss}} \big) & \text{(braking)}
\end{cases}
\end{align}
where the driving and braking conditions are treated separately. During traction, the equivalent mass accounts for the rotational inertia of the wheels, whereas during braking the total longitudinal force is distributed between the front and rear axles according to the brake bias coefficient \(\beta_b\).

\section{Validation} \label{sec:simulation}

\subsection{Online Calibration}
\begin{figure}[tb]
    \centering
    \includegraphics[width=\linewidth]{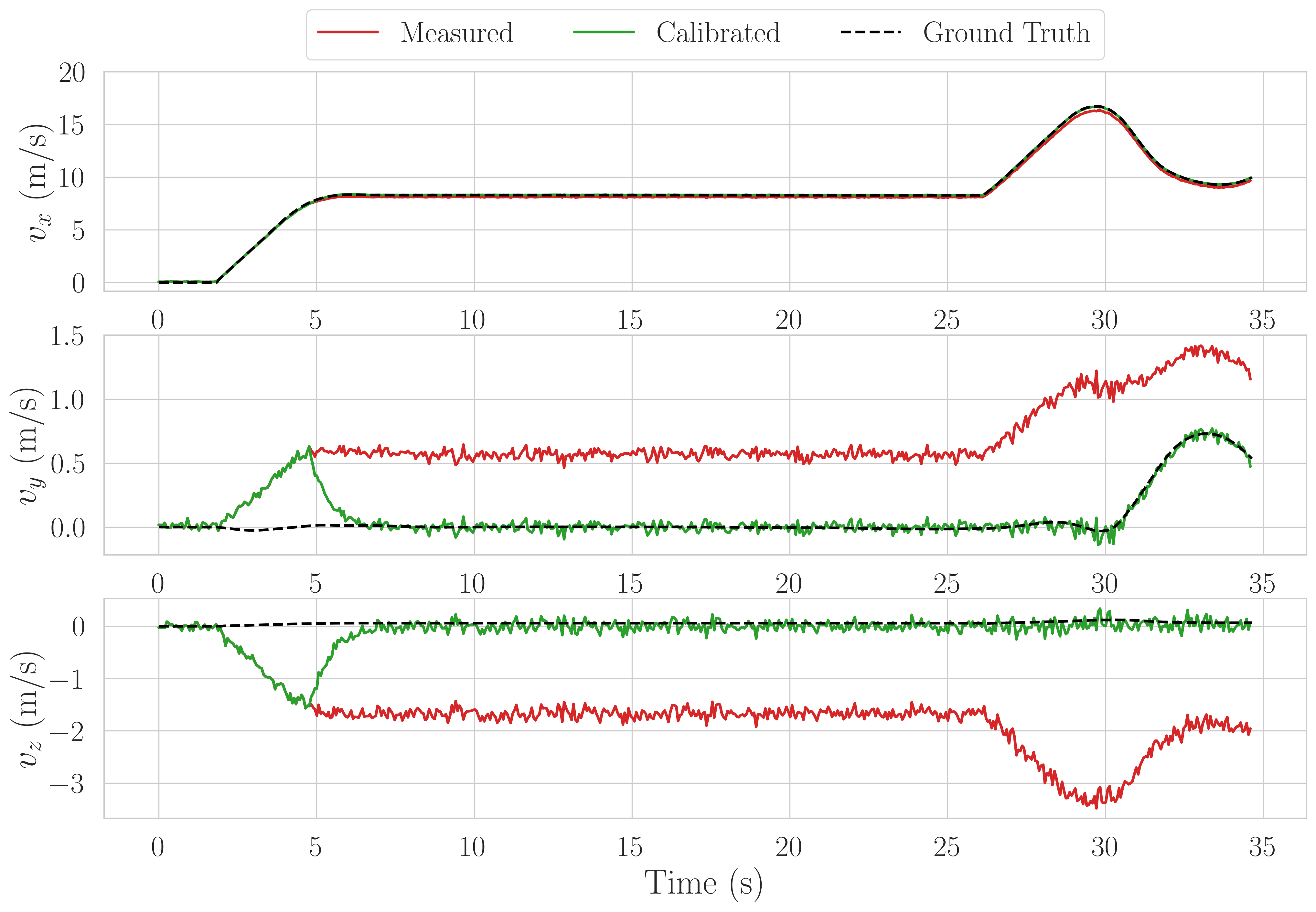}
    \caption{Simulated results of the online auto-calibration method during the initial straight-line driving segment. The \textcolor{red}{measured} estimate is biased due to incorrect rotation calibration, whereas the \textcolor{Green}{calibrated} estimate rapidly converges to the ground truth value.}
    \label{fig:autocalib}
\end{figure}

The proposed online calibration method is evaluated in simulation. In this experiment, the estimator assumes a perfectly aligned RADAR mounting, whereas the generated measurements correspond to a sensor affected by a misalignment of \(15^\circ\) in pitch and \(5^\circ\) in yaw. 
As shown in Fig.~\ref{fig:autocalib}, the vehicle velocity obtained only using RADAR detections exhibits a noticeable bias with respect to the ground-truth values due to the incorrect assumption on the relative rotation between the RADAR and vehicle frames. After approximately \(5\) seconds, the vehicle enters a straight-line driving phase and the online calibration procedure is activated. Following this, the calibrated velocity measurements rapidly converge to the true vehicle velocity.
These results validate the effectiveness of the proposed approach, demonstrating its capability to recover rotation misalignment online during straight driving segments and to significantly reduce the velocity estimation bias induced by incorrect RADAR-to-vehicle frame calibration.

\section{Conclusions and Future Work} \label{sec:conclusion}
Building upon the original RAGE framework \cite{RAGE}, this work presented an extended formulation that jointly estimates the full vehicle state together with lateral and longitudinal tire forces in real time. In addition, an online RADAR calibration procedure was introduced to compensate for sensor misalignment, demonstrating improved accuracy and consistency in velocity estimation in simulation.

Future work will focus on the validation of longitudinal force estimation, both in high-fidelity simulations and on real-world experimental data.

\bibliographystyle{ieeetr} 
\bibliography{sections/refs} 

\end{document}